\documentclass[runningheads]{llncs}
\usepackage{amssymb}
\usepackage{graphicx}
\usepackage{multirow}  
\usepackage{ulem}
\usepackage[figurename=Figure]{caption}
\usepackage{xcolor}
\usepackage{soul}

\usepackage{xfrac} 

\begin{document}
\title{kidsNARRATE: A Versatile Corpus for Studying Chinese-English Bilingual L2 Narrative Skills in Preschoolers 
}
\titlerunning{Non-native Speech Corpus of Bilingual Children}

\author{Hiuching Hung, Andreas Maier, Paula Andrea Pérez-Toro,Tomás Arias Vergara, Sebastian Gesemann, Elmar Nöth, Thorsten Piske}
\authorrunning{H. Hung et al.}

\institute{FAU}
\maketitle              
\begin{abstract}

Narrative skills are important for young children, not only because they are strong indicators of literacy and academic performance, but also because they serve as effective tools to promote children's relationships with the world. However, the linguistic resources for narratives produced by bilingual children are often limited, posing major challenges in the realms of language teaching and language resource studies. Moreover, with the increasing prevalence of remote data collection, few guidelines are available for collecting such data remotely. In this paper, we present kidsNARRATE. KidsNARRATE is a non-native speech corpus designed to study the narrative comprehension of Chinese-English bilingual children in their L2 English. KidsNARRATE comprises 6 hours of audio recordings of children taking the narrative test Multilingual Instrument for Narratives (MAIN), as well as the manual transcription, human-rated scores, and annotations of grammatical and pronunciation errors at the word level. The audio recordings of the English section have been processed to meet the requirements of certain machine learning applications. For reference, we audio- and video recorded the same children taking the parallel MAIN test in L1 Chinese. 

In the course of this study, we developed a remote recording method. By using accessible recording tools such as ZOOM and Open Broadcaster Software (OBS), this easy-to-use method can serve as a specific template for researchers and teachers seeking to remotely record audio and/or video data for linguistic studies. In general, the rich linguistic content and the compatibility with machine learning processes make kidsNARRATE a valuable resource for studies of early child L2 acquisition and automatic speech recognition (ASR). Lastly, future work regarding data collection methods and second language teaching are proposed.   

\end{abstract}

\section{Introduction}
\subsection{Background}
Narrative skills play a fundamental role in children's language development. Starting in infancy, children build connections to the world by receiving narrations about their everyday experiences, most often told by their caregivers. At older ages, children start using their first words to express needs and feelings, which can be seen as an early form of narrative \cite{paris2003assessing,stein1997goal}.The study of narratives, as Connelly and Clandinin aptly state it, helps us gain insight into how humans experience and make sense of the world \cite{connelly1990stories}. From an educational perspective, narrative competency has been identified as a strong indicator of academic performance in school-aged children, particularly literacy \cite{suggate2018infancy,bigozzi2016tell,reese2010children}. However, questions about the development of narrative competence in the bilingual child population remain unanswered. Researchers worldwide have reported that, besides the asymmetric language development and the diverse language parings of children's L1 and L2, one critical issue is the scarcity of language learning resources for this specific group \cite{newbury2020current,knudsen2021multilingual,altman2016macrostructure}. Delavan, Freire and Menken report that the current language of instruction in the U.S. school programs has exclusively been English, even though the vast majority of multilingual learners attend these programs. The home languages of the learners are neither integrated into the educational system, nor receive real support from language education policies \cite{delavan2021editorial,baker2011foundations}. Despite the continued development of Two-Way Immersion (TWI) over the past half-century, and its well-documented benefits, the advancement of TWI still faces a number of challenges due to constraints such as the shortage of qualified bilingual teachers\cite{de2016two,kim2015bilingual}. This prevailing disregard for linguistic diversity and the resulting lack of resources not only impairs research efforts, but also severely diminishes the chance for bilingual integration among policymakers.

Constructing a speech corpus might present a direct solution to the aforementioned issues, especially when supported by the rapid advances in speech technologies \cite{golonka2014technologies,kannan2018new}. Nowadays, ASR-based technologies have been widely integrated into various learning scenarios, helping teachers design learning activities, assess language skills, and provide tailored feedback to learners \cite{golonka2014technologies,neri2003automatic}. On the other hand, the success of ASR development is strongly tied to the training dataset. For example, the word error rate (WER) of ASR applied to children's speech is typically two to five times worse than that with adult speech \cite{shahnawazuddin2017effect}. 

During the Covid-19 period, collecting data via the internet quickly surged to be the next preferred solution. However, this does not imply that collecting children's speech data is any less difficult, if not more so. Younger children cannot read and have shorter attention spans \cite{hynninen2017multilingualism,oliver2005constraints}, also, they are often unfamiliar with wearing headsets, which is commonly required for online data collection, and may exhibit less cooperation during the process. Linguistically, children’s speech often contains ungrammatical or ambiguous input, including incomplete sentences, grammatical errors, and unintelligible words and phrases that are difficult to process \cite{radha2022audio}.Furthermore, remote data collection is highly dependent on participants' personal equipment, setup, and recording environment, all of which introduce variability in audio quality. These factors require careful consideration of data recording constellations to mitigate inconsistencies in audio quality, and to prevent inaccurate interpretations of segment and voice quality contrasts that could ultimately compromise the results \cite{ge2021reliable}. To date, there has been little comprehensive guidance on online data collection methods, and even less specifically aimed at the remote collection of speech data from young children.
  
The contribution of this study is twofold. Firstly, we create a speech corpus kidsNARRATE, a non-native speech corpus designed for the studies of the narrative comprehension of Chinese-English bilingual children in L2 English. We plan to make the audio section publicly available upon request. Secondly, we propose a remote data collection method that can be applied to collect speech data remotely in similar linguistic studies.

\subsection{Diverse speech corpora and innovative data collection methods}
According to a 2022 review paper by Sobti, Kadyan and Gluleria, worldwide only twenty children's speech databases are publicly accessible to researchers \cite{sobti2022challenges}. While the availability of children's speech corpora remains limited, interest in exploring and expanding children's speech corpora have been growing. Recent trends in corpus development show that corpora nowadays are often created for specific purposes and, accordingly, have distinctive features. We now present some datasets that exemplify this initiative.One notable example is the Child Language Data Exchange System (CHILDES). Established in 1984, the CHILDES has emerged as a large-scale children’s speech corpus that enables researchers to search, process, share and exchange contents (transcriptions, audio and video). To ensure consistency across transcriptions, CHILDES works compatibly with CHAT, a standard transcription format for transcription as well as phonological and morphological analysis. In the present study, we also used the CHAT format for our transcriptions.

In the realm of non-native speech databases, Speechocean762 is a newly launched, open-source speech corpus designed for pronunciation assessment. It contains 5000 English utterances from 250 non-native speakers, half of whom are children. Speechocean762 is manually annotated at the phonemic, word and sentence level, with a baseline system for download and other use \cite{zhang2021speechocean762}. Another non-native example is the TLT-school speech corpus. With a primary focus on the assessment of children's L2 proficiency, TLT-school includes the speech data from 9- to 16-year-old Italian students who learn both German and English at different levels of proficiency \cite{gretter2020tlt}. A large portion of the corpus has transcriptions and all utterances are manually scored. Due to the wide age range and the language level classification, the TLT-school serves as a valuable resource for non-native speech recognition and automatic assessment of L2 proficiency. The English section of TLT-school was employed in the ASR challenge of non-native children’s speech at Interspeech 2021. 

SingaKids-Mandarin is a non-English speech corpus specifically designed to study the phonetic characteristics of children in Singapore. It consists of 125 hours of Mandarin reading audio recordings of 7- to 12-year-old Singaporean children. This corpus includes rich phonetic annotations such as phonetic transcriptions, lexical tone markings, and proficiency scores at the utterance level. Researchers have used this corpus to identify distinctive pronunciation patterns of the Singaporean children and some related topics \cite{chen2016singakids}.

With its pioneering suitability for machine learning applications, the speech corpus kidsTALC has made a special contribution to the field of language assessment. KidsTALC is the first of its kind to fulfill all the requirements for automatic language assessment. This corpus provides 25 hours of high quality continuous speech from three to eleven year old German children, accompanied by manual, revised, orthographic and phonetic transcriptions \cite{rumberg2022kidstalc}. However, both TST-school and kidsTALC have relatively large age spans of seven and eight years, respectively. While providing valuable insights, children's speech data with extended time span might lead to a degradation of ASR performance, due to the significant physical and linguistic changes in children's speech that come with age \cite{wilpon1996study}. 

Recent trends also demonstrate an expanding interest in the fusion of different datasets. Batliner and his colleagues conducted a cross-linguistic analysis of verbal and emotional interactions between the Sony's AIBO robot and the German and English children in respect to language, emotional speech, human-robot communication, as well as read vs. spontaneous speech. Their study provides the multifaceted insights into human-robot interaction and cross-linguistic comparisons drawn under different experimental conditions \cite{batliner2004you}. MERLIon is a newly launched speech corpus with more than 30 hours of video call datasets recorded via ZOOM. The videos present the parent-child book-reading sessions with the spontaneous and naturalistic speech. Particularly, the audio recordings of MERLIon include examples of English-Mandarin code-switched speech and English in non-standard accents, making MERLIon a unique data resource for advancing the reliability and robustness of language identification and language diarization \cite{chua2023merlion}.

To accommodate the dynamic societal settings and the diverse research needs, researchers have explored proactively innovative methods to collect data. Recent trends in data collection involves exploring and assessing the efficacy of readily available recording devices such as smartphones and online platforms. Ge, Xiong and Mok compared the acoustic performance of various combinations of recording devices and environments. Their findings indicated that the remote recordings made with smartphones and ZOOM can capture useful information for prosody research, but should be used with consideration for segmentation tasks \cite{ge2021reliable}. Archibald et al. examined the experience of 16 nurses who participated in online interviews via ZOOM. The results showed a high level of participant acceptance when using ZOOM for interview, suggesting its viability as a tool for qualitative data collection \cite{archibald2019using}. In line with the work of Archibald and her colleagues, Leemann et al. investigated the potential differences between in-person interview and visual settings. The data with visual setting come from the experience of 36 participants using ZOOM and smartphones. The results demonstrated a high level of consistency between these two data collection methods, indicating that the online data collection method could yield reliable results for linguistic work \cite{leemann2020linguistic}. 

\section{Key features of kidsNARRATE}

In this section, we outline the distinctive features that characterize kidsNARRATE: 
\begin{itemize}
    \item[$\bullet$] \textbf{Standardized tests with human-rated scores as ground-truth} serve as a benchmark for the automatic assessment of the language proficiency.
    \item[$\bullet$] \textbf{Transcriptions annotated with word-level grammatical and pronunciation errors} provides valuable linguistic resources, such as language patterns, for a wide range of research areas.
    \item[$\bullet$] \textbf{Accompanying videos} help enhance the transcription accuracy. In some ambiguous situations,  the paralinguistic cues that the video recordings provide  can be used for studies that involve facial emotion recognition.
    
    \item[$\bullet$] \textbf{Our audio data is processed with audio signal processing technologies such as noise reduction and trimming, to meet the requirements of machine learning.}    
    \item[$\bullet$] \textbf{Participants with a narrow age range} are beneficial for extracting targeted, age-related speech features for the enhancement of ASR performance.  
    \item[$\bullet$] \textbf{Parallel tests in L1 Chinese} are provided as the baseline for potential language transfer analysis.   

\end{itemize}
This paper is organized as follows. Section \ref{sec:datacollection} describes the data collection process. Section \ref{sec:dataprocessing} focuses on the audio data processing, transcription and annotations. Section \ref{sec:useofdata} discusses some possible uses of this corpus, including applications in ASR-related aspects and second language teaching. Finally, direction in future work concludes this paper. 
\section{Data collection}
\label{sec:datacollection}
\subsection{Preparations}
\label{sec:preparations}

In response to the COVID-19 pandemic and China's zero-COVID policy, we conducted our study remotely with the accessible recording tools. During this period, it was rather difficult to find a suitable kindergarten to be our research partner, as most Chinese kindergartens were fully occupied with the consequences of the pandemic. To overcome this problem, we emphasized the potential of the study in assessing children's language skills and teaching quality, both of which appeal to the pedagogical goals of most kindergartens. After an extensive six-month search, we established a collaborative relationship with a Chinese-English bilingual kindergarten in Quanzhou, Fujian Province, China. 

\subsubsection{Teacher training and parental consent form}
A group of five teachers received one-hour online training on the MAIN administration protocols from the researcher \cite{bohnacker2019background}. One teacher was selected as the coordinator and is responsible for organizing communication with the researcher. Another teacher with a background in computer science in team provided technical support during all recording sessions. An online group was created to facilitate communication among the teachers. Although all teachers have functional digital literacy and the skills to use computers for teaching, none of them had previous experience with online research or had used MAIN before.
All the parents and teachers signed the consent forms to participate in this study. 

\subsection{Technical setup}
\label{sec:techsetup}
\subsubsection{Recording procedure}

Each participating child performed two MAIN tests in one recording session — one in English and one in Chinese — resulting in two videos for two perspectives (child \& teacher).
During the English tests, each test lasted approximately 7 to 11 minutes, with about 3 to 5 minutes for storytelling and warm-up conversation, and the remaining time for answering the questions. Children's age and score distributions are given in Table 2. 
Three raters were present during each session: two teachers were physically present in the room, while the third rater, who was the researcher located in Germany, participated via ZOOM. All teachers' voices are female only. The male teacher participated as a silent rater. Each rater gave scores independently, and any disagreements about the scores were discussed until resolved. The Cronbach's alpha for inter-rater agreement in English = 0,977, inter-rater in Chinese = 0,968, indicating a high level of agreement among the raters.
 \begin{figure}[htp]
    \centering
    \includegraphics[width=6cm]{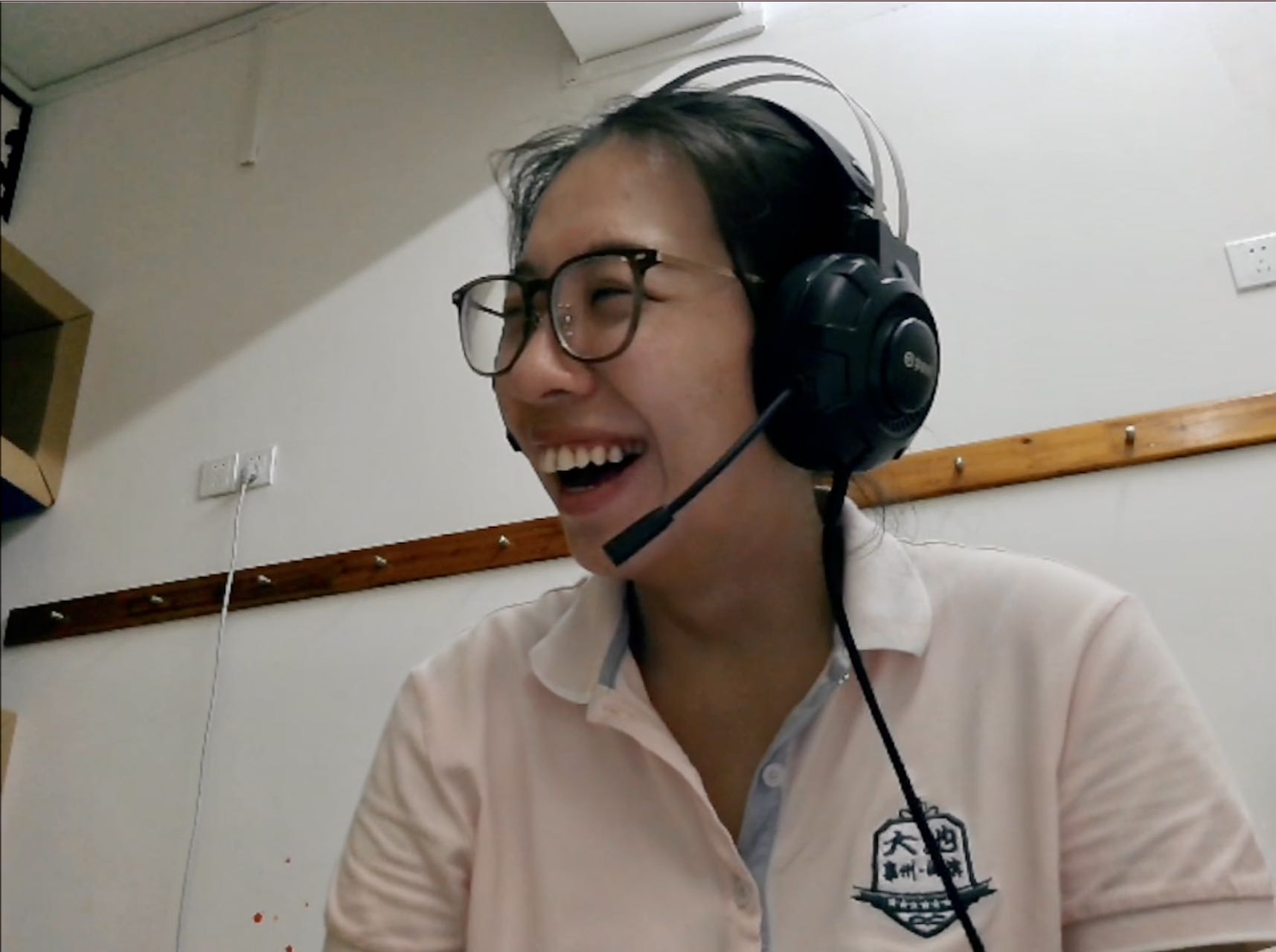}
    \caption{Example of teacher camera perspective}
    \label{figure:teachercam}
\end{figure}

\begin{figure}[htp]
    \centering
    \includegraphics[width=6cm]{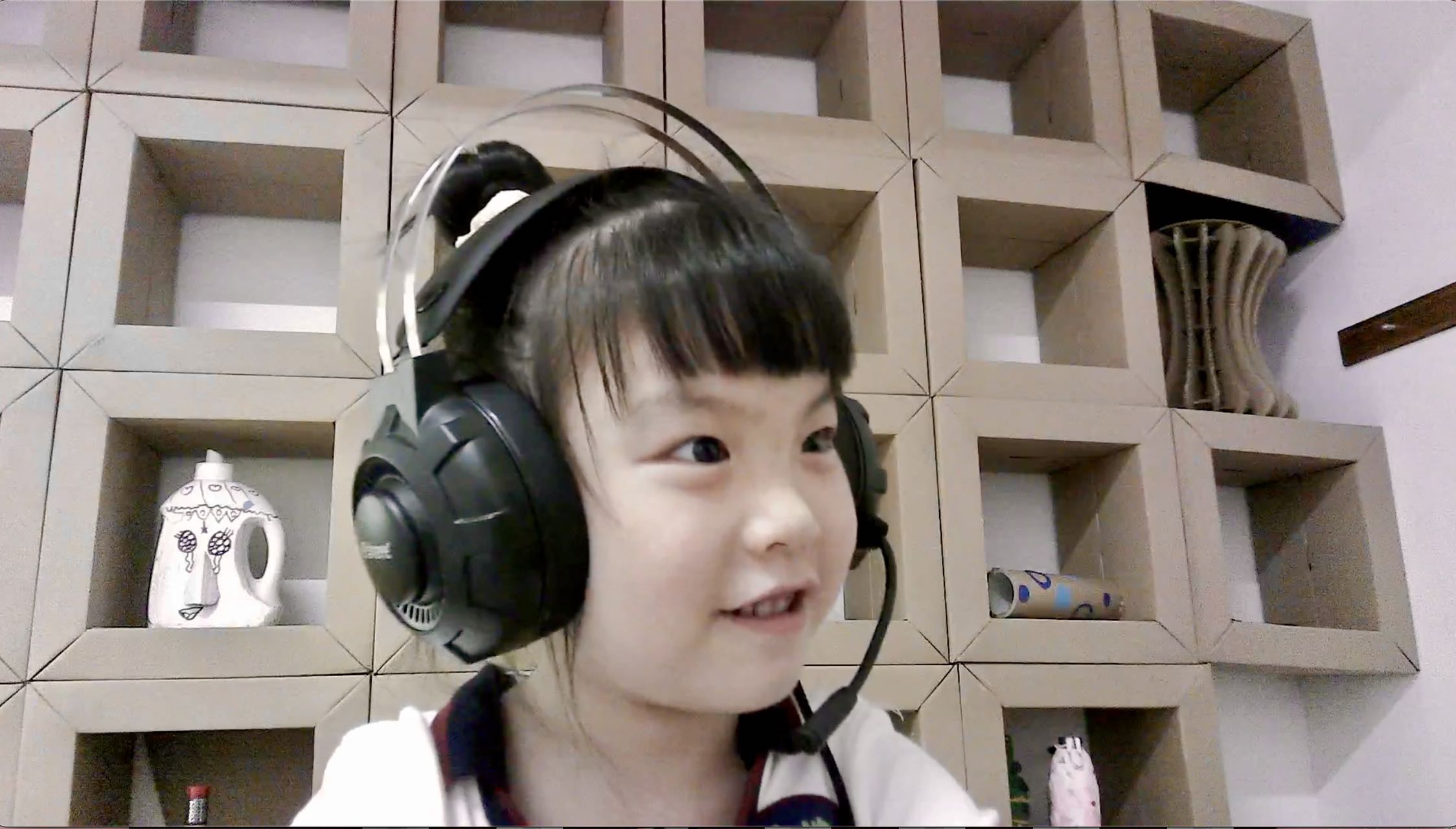}
    \caption{Example of child camera perspective}
    \label{figure:childcam}
\end{figure}

\subsubsection{Equipment}
All data were recorded with the same equipment in the teacher's office during June and July, 2022. We chose this particular room because there were two large bookshelves facing each other that could help absorb noise. Consider the accessibility of recording tools, we used a constellation of accessible recording tools, including ZOOM, OBS, and the wired HD headsets with built-in microphone to enhance the audio quality. ZOOM and OBS Studio both function well in China. Figures 1 and 2 show the camera perspectives of the teacher and child. The audio was recorded at a sampling rate of 48 kHz, and the video was recorded at 720p using an external webcam connected to a laptop. One main purpose of recording the video is to provide an additional source for transcription and annotation in case of poor audio conditions. Two computers were configured with the audio and video recording setup to record children and teachers, respectively. As shown in Figure 3, the children and teachers were seated facing each other, and the microphones were carefully placed near the mouths to avoid exhalation sounds while minimizing potential background noise. In addition, the researcher in Germany participated throughout the tests via ZOOM. 

\begin{figure}[htp]
    \centering
    \includegraphics[width=6cm]{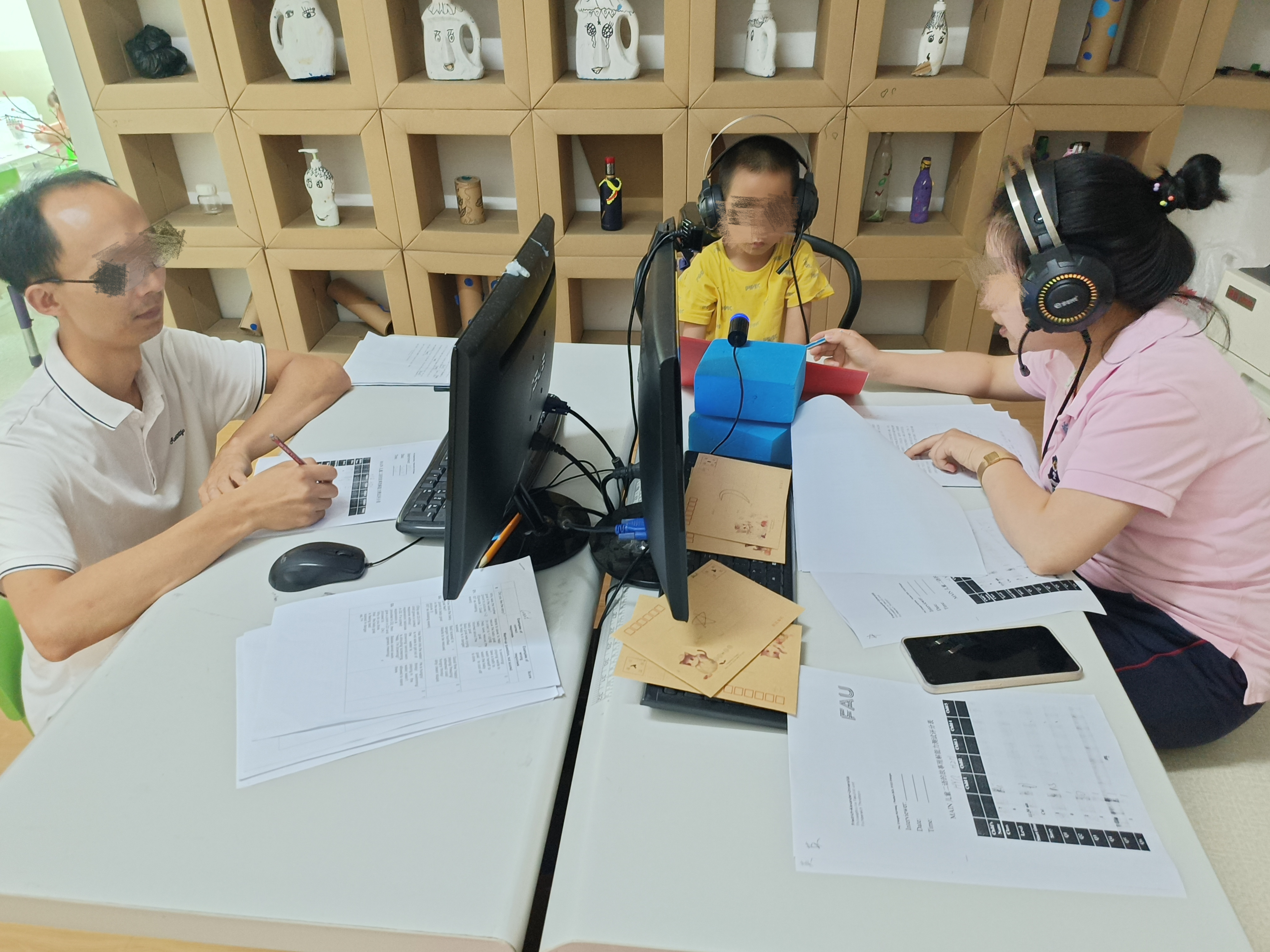}
    \caption{Recording positions}
    \label{fig:galaxy}
\end{figure}

\subsection{Participants}
\label{sec:participants}

Fifty children (26 boys and 24 girls) aged five to seven years old (mean age = 5 years, 10 months [5;8], SD = 6 months, range =5;0-6;8), five kindergarten teachers (4 female, 1 male), and one researcher living in Germany participated in this study. All participants are Chinese native speakers. Some children speak the regional Min-Nan dialect at home. No hearing or cognitive problems were reported in the questionnaires completed by the parents. Most of the children had their first exposure to English in kindergarten.

\setlength{\tabcolsep}{6pt}
\begin{table}
    \centering
    \begin{tabular}{cccccc}
        \hline
        Age & Mean & Standard & \multicolumn{2}{c}{Number of students} \\
        (months) & (MAIN) & Deviation & MS $\geq 5$ & MS $< 5$ \\
        \hline
        60-72 & 3.889 & 2.561 & 14 & 22 \\
        \hline
        73-84 & 3.786 & 2.577 & 7 & 7 \\
        \hline
        \vspace{1mm}
    \end{tabular}
    \caption{ MAIN scores (MS) distribution from different age groups}
    \label{tab:maindistribution}
\end{table}

\subsection{Stimuli}
\label{sec:materials}

The MAIN test was used in this study \cite{gagarina2019main}. MAIN is a test designed to assess the narrative skills of bilingual children. The test works as follows: the child chooses one out of four modal stories provided by MAIN. Each modal story is accompanied by a sequence of six wordless pictures that the child can view throughout the tests. The teacher narrates the chosen story, then asks ten standard questions that are all related to the story and also provided by MAIN, for the child to answer. Each correct answer earns one point. The points range is from 0 to 10.

We believe that the MAIN test is suitable for the following reasons: First, the MAIN test has been extensively tested worldwide for over a decade for assessing the narrative abilities of bilingual children ages 3 to 10; Second, it has a user-friendly design that includes protocols, scoring sheets and examples of correct responses, promoting comparable results across languages; Third, MAIN's wordless pictures are familiar and appropriate for preschool children, facilitating elicitation.

\section{Data processing}
\label{sec:dataprocessing}

\subsection{Audio data processing}
\label{sec:audiodataprocessing}

As mentioned above, each recording session has two videos, one recording the teacher and the other one the child. Both videos contain the L1 Chinese and L2 English test. 

\begin{figure}
    \centering
    \includegraphics[width=0.7\textwidth]{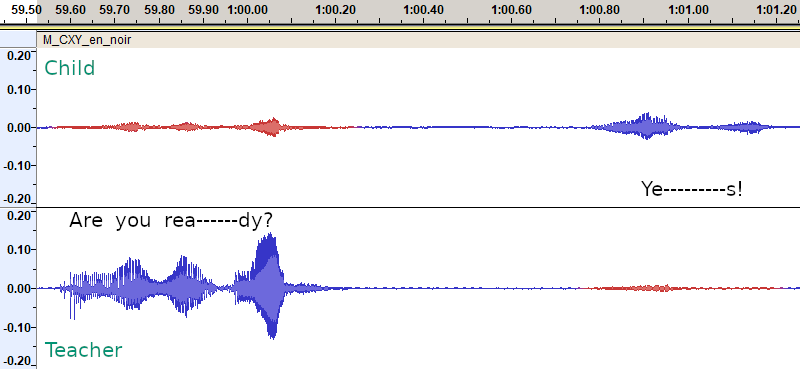}
    \caption{Example waveform view of the child and teacher audio channels that contain crosstalk: both
microphones picked up both speech signals. Shown in blue is the wanted speech and in red the unwanted crosstalk.}
    \label{fig:xtalk1}
\end{figure}

We separated the English and the Chinese videos and converted each of them into the WAVE audio file format (.WAV). This is done to prepare for the noise and crosstalk reduction. Due to the two microphones and two speakers in the same room, both speakers are audible in both recordings. This is called crosstalk in multichannel audio data \cite{wrigley2004speech}. To eliminate crosstalk is essential to ensure good quality training data for modeling. The figure \ref{fig:xtalk1} shows an example in this case. 

 Several technical methods and algorithms have been proposed to separate and cancel crosstalk between multiple speakers in the audio data. Maier, Haderlein, and Nöth proposed an innovative approach to adapt speaker-independent recognisers, trained on audio recorded collected from 33 speakers with close-talking microphones, to a speaker-open environment. This adaptation approach significantly improved performance in distant speech conditions \cite{maier2006environmental}. To address the crosstalk interference in the multichannel close-talk audio scenarios, Meyer, Elshamy, and Fingscheidt introduced an adaptive filtering method that was originally used for acoustic echo cancellation (AEC) to mitigate microphone leakage. Their results improved the speech-to-noise ratio by up to 2.7 dB. Second, they also investigated why AEC algorithms can be used to estimate interfering signals and room impulse responses, which typically require clean reference channels without crosstalk \cite{meyer2020multichannel}. Our approach employed an adaptive noise gate effect that can discriminate between wanted speech and unwanted crosstalks. This is achieved by dynamically deriving speech activity thresholds based on short-term averaged powers of both signals under the assumption that the crosstalks are present. For example, a high power segment of the teacher's audio recording would raise the child's speech activity threshold and vice versa. This will suppress even stronger crosstalk while maintaining sensitivity to the desired speech signals. The software used is Audacity (version 3.2.5).

\subsection{Transcriptions and annotations}
\label{sec:transcriptionsannotations}

\subsubsection{Transcription}


We transcribed our audio data using a combination of a selected ASR service and manual review. We selected our ASR service as follows: first, we selected two popular commercial ASR services, Amazon Web Services (AWS) and Google ASR, and three random, one-minute audio passages from the recordings. A Chinese PhD student with a master's degree in linguistics has carefully transcribed these three passages for later comparison. To ensure an unbiased and objective comparison, we intentionally introduced a three-week interval between the initial manual transcription and the subsequent ASR-based transcriptions. The rationale behind this approach was to allow the transcriber's memory of the initial manual transcription to naturally fade over time, and ultimately to minimize any potential memory bias on the account of the skilled transcriber. After the three-week interval, we used AWS and Azure service to transcribe the same three audio passages and compared the ASR-generated transcriptions to the manual transcriptions. As shown in Table 2, the AWS yielded an average WER of 7.7\% and Google ASR 17.3\%. We then used AWS as our ASR service for this study.

\setlength{\tabcolsep}{6pt}
\begin{table}
    \centering
    \begin{tabular}{cccccc}
        \hline
          & Sample 1   & Sample 2   & Sample 3   &   \\
        ASR & \ ``CYS''        & \ ``FYX''        & \ ``WKT''        &  Average  \\
        Methods  & (79 words) & (62 words) & (89 words) & (77 words) \\
        \hline
        & 4 errors & 5 errors & 9 errors &     \\
        AWS & WER 5.1\% & WER 8.1\% & WER 10\% & WER 7.7\%  \\
        \hline
        Google Web & 15 errors & 12 errors & 14 errors &           \\
        Services   & WER 19\%  & WER 19\%  & WER 16\% & WER 18\% \\
        \hline
        \vspace{1mm}
    \end{tabular}
    \caption{Comparing WER of two ASR services. The samples CYS, FYX and WKT each presents a children recordings}
    \label{tab:wer}
\end{table}

We used the AWS service to process each audio files into JSON files, which is a machine-readable text format. These files contain lists of ``segments'' that roughly correspond to utterances. Within each segment, the files contain a start and an end time and the corresponding transcribed text. Using a custom Python script, we converted these transcripts from JSON to the CHAT format, individually adapted for each speaker \cite{macwhinney2000childes}. This approach treated each JSON segment as a distinct utterance and converted it into a line in the CHAT file with speaker ID, text and time code range. In the transcription, each sentence is marked with two time stamps that correspond to the beginning and end of the sentence. 

We are aware that it is not always possible to have access to corresponding videos for every transcription task; however, the advantages of transcribing with videos are significant. Visual images make it easier to identify speakers and clear the ambiguities. It is also easier to deal with incomplete sentences, grammatical errors and mispronounced words, as well as noise such as breath, coughing and laughing, which are often unavoidable in the recordings. In situations where we can not clearly identify children's speech, we consult with the teachers or parents who are familiar with these children. The transcriptions were further double-checked by two senior kindergarten teachers for verification. Additionally, we also provide an orthographic transcription of the Chinese test recordings.

\subsubsection{Annotations} 
All audio recordings are manually annotated with grammatical and mispronunciation errors at the word level, following the CHAT coding guidelines. For instance, \textit{the bird feel [: feels] [*] angry} is marked as a grammatical error; \textit{the dog noticed the boy’s bag and sought [: thought] [*]} is marked as a pronunciation error. This is done to prepare the corpus for future public availability on the CHILDES platform. Two approaches are used to ensure the credibility and the accuracy of the annotations. First, to make sure that we translate the children's utterances correctly, we referred to the video recordings for clarity. Another solution is to consult with the teachers that are familiar with children's English. This annotation process was supported by two experienced kindergarten teachers.  

It is worth noting that the errors marked in the corpus also include those made by teachers. Therefore, future work can further categorize the pronunciation errors, and investigate how teachers' errors affect children's errors. 



\section{Use of the data}
\label{sec:useofdata}
We are in the process of finalizing kidsNARRATE soon. Once completed, we plan to make the audio recordings and the transcriptions publicly available upon request. Meanwhile, the preliminary results have been used in a study examining the relationship between children's intelligibility and their narrative comprehension in L2. Here are some suggestions for using this corpus: 
\begin{itemize}
     \item[$\bullet$] \textbf{ASR-related topic}: kidsNARRATE has great potential for ASR-related development, particularly in the realm of children's speech recognition. The detailed acoustic data available in kidsNARRATE, including pronunciation, prosody, speech rhythm and syntax, enhances its utility for developing the L2 learning tools customized for children. Further, the distinctive narrative nature of kidsNarrate can be leveraged in the design of gamification activities for a more interactive engagement experience.
     
    \item[$\bullet$] \textbf{Pedagogical implications}: 
 The impact of teachers' L2 input on children's long-term perception and production has always been an area of focus. As Piske points out, children can only develop a native-like or even native L2 skills if they receive a significant amount of native or native-like pronunciation and grammatically correct input from teachers. Therefore, teachers should be mindful of their own challenges with articulation and language perception, as their awareness contributes (sometimes even unconsciously) significantly to providing the language models for the learners \cite{piske2008input,piske2010small}. kidsNARRATE annotates the pronunciation and grammar errors made by both the teachers and the children, facilitating the studies of the similarities between them. 
 
 \item[$\bullet$] \textbf{Automated assessment of language proficiency}: One of the main goals of this corpus is to provide data resources for the development of automated assessment of children's L2 narrative comprehension. Traditionally, L2 narrative comprehension has been difficult to measure, mainly due to the lack of appropriate measures and the labor-intensive nature of human ratings. An automated approach to the L2 narrative comprehension assessment would not only be cost-effective, but would also help to alleviate the burdens on teachers.  
    
    \item[$\bullet$] \textbf{Facial emotion recognition}: the video recordings of kidsNARRATE provides potential for extracting paralinguistic cues, such as observing and analysing emotions through gesture, laughing, facial expressions, and overall interactions between teachers and children. These additional layers of information enrich this dataset with more insights into the emotional dimensions for a comprehensive analysis of children's L2 performance. 
    
\end{itemize}

\section{Conclusion and future work}
\label{sec:conclusion}

The rapid convergence of speech science and speech technologies has been prompting researchers to create language resources such as tailored language corpora for various, specific tasks. To address the growing demand for more children's speech corpora in the realm of natural language processing, this paper presents the development and compilation of kidsNARRATE, a non-native children's speech corpus with rich annotations and versatile usability. Moreover, we have developed an innovative remote data collection method using accessible recording devices. This method not only supports our study, but also provides a specific template that can collect high-quality speech data remotely for linguistic purposes.  
 
Future research can be considered from three aspects. Firstly, to further explore the feasibility and suitability of the remote data collection method, we encourage researchers to test the method described in this study for various purposes and across diverse settings, for example, exploring its effectiveness in collecting pathological speech or the speech produced by younger groups (for younger groups, we could consider using the wireless collar microphone). Simultaneously, issues such as ensuring data security, addressing ethical considerations and refining remote data collection procedures should be discussed.

Secondly, researchers could expand the scope of linguistic annotations and build acoustic models using the kidsNARRATE corpus. Subsequently, our next focus is to leverage the video recordings from kidsNARRATE to explore the relationship between children's L2 narrative comprehension and child-teacher interactions, particularly in relation to emotional speech analysis that involves facial expressions. To deepen the understanding of the multifaceted interactions between language and emotion in the learning context could help us gain a holistic pedagogical perspective.

Finally, it should be noted that kidsNARRATE was collected during COVID-19 pandemic. We posit that this unique context may have contributed to a potential learning deficit among children, with the immersive learning concept offered by the kindergarten not reaching its full potential. In upcoming studies, we aim to investigate any potential differences in children's language skills under different settings. In addition, we are also interested in exploring innovative elicitation techniques used by teachers to encourage speech output from children while ensuring that they feel comfortable and engaged during learning.



  
%
%
%
%
\bibliographystyle{splncs04}
\bibliography{paper}

\end{document}